\documentclass[a4paper]{article}

\usepackage{INTERSPEECH2018}
\usepackage{times}
\usepackage{latexsym}
\usepackage{color}
\usepackage{tabularx}
\usepackage{multirow}
\usepackage{caption}
\usepackage{subcaption}
\usepackage{tikz}
\usepackage{tikz-qtree}
\usepackage{url}
\usepackage[utf8]{inputenc}
\usepackage{xcolor}
\usepackage{pgfplots}
\pgfplotsset{compat=1.14}

\definecolor{bblue}{HTML}{4F81BD}
\definecolor{rred}{HTML}{C0504D}
\definecolor{ggreen}{HTML}{9BBB11}
\definecolor{darkerggreen}{HTML}{9BBB88}
\definecolor{ppurple}{HTML}{9F4C7C}
\definecolor{yyellow}{HTML}{FAF532}
\makeatletter
\def\url@leostyle{%
  \@ifundefined{selectfont}{\def\UrlFont{\sf}}{\def\UrlFont{\small\ttfamily}}}
\makeatother
\newcolumntype{R}{>{\raggedleft\arraybackslash}X}

\title{Coherence Models for Dialogue}
\name{Alessandra Cervone, Evgeny A. Stepanov, Giuseppe Riccardi}
\address{
Signals and Interactive Systems Lab, 
  DISI, University of Trento, Italy}
\email{\{alessandra.cervone, evgeny.stepanov, giuseppe.riccardi\}@unitn.it}

\begin{document}

\maketitle
\begin{abstract}
Coherence across multiple turns is a major challenge for state-of-the-art dialogue models.
Arguably the most successful approach to automatically learning text coherence is the entity grid, which relies on modelling patterns of distribution of entities across multiple sentences of a text.  Originally applied to the evaluation of automatic summaries and the news genre, among its many extensions, this model has also been successfully used to assess dialogue coherence. 
Nevertheless, both the original grid and its extensions do not model intents, a crucial aspect that has been studied widely in the literature in connection to dialogue structure.
We propose to augment the original grid document representation for dialogue with the intentional structure of the conversation. 
Our models outperform the original grid representation on both text discrimination and insertion, the two main standard tasks for coherence assessment across three different dialogue datasets, confirming that intents play a key role in modelling dialogue coherence.
\end{abstract}

\section{Introduction}
This work addresses the problem of automatic coherence assessment of dialogue.
% * <alessandra.cervone@unitn.it> 2018-06-17T03:14:53.955Z:
%
% ^.
% * <alessandra.cervone@unitn.it> 2018-06-17T03:14:50.831Z:
%
% ^.
Coherence -- what makes a text unified rather than a random group of sentences -- is an essential property to pursue for a system aimed at conversing with humans. 
%Since dialog relies on a fundamental principle of mutual cooperation \cite{grice1970logic} between the interlocutors, if one participant gives incoherent responses the other participants might feel a breach in this cooperation principle.   
Nonetheless, producing coherent responses across conversation turns remains an open research problem for state-of-the-art (SoA) open-domain dialogue models \cite{li2016diversity,li2016deep}.

Furthermore, progresses in open-domain dialogue modelling are currently curbed by a lack of standardized automatic metrics to evaluate and compare conversational systems \cite{liu2016not}. 
Most available automatic metrics for dialogue evaluation either rely on surface features such as the words used (e.g. BLEU \cite{papineni2002bleu}), try to replicate generic human judgments \cite{lowe2016evaluation}, or work only for task-based dialogue systems \cite{walker1997paradise}. For evaluation, the field still relies heavily on user satisfaction, an expensive and time-consuming process which poses its own challenges given the subjectivity of human judgment.
While coherence has been proposed multiple times as an important metric to evaluate open-domain dialogue, there have been only few studies on open-domain dialogue coherence assessment \cite{gandhe2016semi,higashinaka2014evaluating,venkateshevaluating}. %rely on utterance level manual annotation of coherence, a requirement which does not allow to scale. 
% These metrics were never measured in the standard coherence tasks

% Coherence models in NLP represent a formalization of one aspect of the structure of a document: local, unsupervised
On the other hand, the Natural Language Processing (NLP) literature has made several attempts \cite{grosz1995centering,barzilay2008modeling} to formalize the notion of text coherence into \textit{coherence models}. 
The entity grid, the most popular approach to coherence modelling in this community, proposes to represent documents according to the patterns of distribution of entities mentioned in the text across adjacent sentences \cite{barzilay2008modeling}. 
Besides its correlations with human judgment, among the reasons behind the success of this approach is the fact that it is linguistically motivated, capturing important aspects of discourse coherence related to entities distribution \cite{joshi1979centered,givon1987beyond}. 
% TODO: Add here something more specific about Focus and Topic
Since its original proposal, the entity grid has undergone multiple extensions and has been widely applied to different tasks such as text coherence rating, automatic summaries assessment and sentence ordering, among others \cite{barzilay2008modeling,elsner2011extending}. It has also been successfully applied to dialogue \cite{purandare2008analyzing,elsner2011disentangling}, for example for chat disentanglement. 

Being a local coherence model, i.e. modelling paragraphs internal coherence rather than the global coherence of the entire text, the extensions of the grid proposed for dialogue do not take into account one essential characteristic of dialogue coherence that has been studied for several years: its intentional structure.

% Intentional structure
Several theories studying dialogue coherence are indeed rooted on the idea of an internal structure given by participants' intents in a conversation \cite{sacks1974simplest,sacks1995lectures,schegloff1968sequencing,schegloff1973opening}. 
In many approaches, the basic units of these sequences are a variation of Dialogue Acts (DAs), a concept based on Speech Acts theory \cite{austin1975things}, that conveys the illocutionary function of an utterance in a conversation; and represents a formalized and generalized lexicon of speaker intents. 
Attempts to formalize computationally similar theoretical intuitions about dialogue coherence \cite{grosz1986attention,allen1980analyzing} did not find wide-spread application, since they require extensive expertise and significant manual annotation effort.
%Approaches that tried to formalize computationally similar theoretical intuitions about dialogue coherence \cite{grosz1986attention,allen1980analyzing} found limited real-world applications, since they require significant manual effort and extensive expertise.
% rely on manually annotated data and extensive expert-design to be applied to new domains.

%TODO: how our work is different from Higashinaka and David Traum!
% This proposal: coherence models for dialog
We propose entity-grid inspired coherence models for dialogue augmented with intentional information, 
%in our approach 
represented by DA transitions across turns.
%In this work, we propose entity-grid inspired coherence models for dialogue augmented with information about DA transitions across turns (thus conveying the intentional structure). 
To the best of our knowledge, this work is the first to combine entity grid coherence models with DAs. 
We compare our models to the original entity grid on the two de-facto standard tasks for coherence, i.e. sentence (in our case turn) ordering discrimination and insertion.
We perform our experiments on three publicly available datasets conveying different types of dialogue (task-based and open-domain) and DAs annotation schemes, namely Switchboard \cite{godfrey1992switchboard}, AMI \cite{amicorpus} and Oasis \cite{leech2003generic}. Our results show the crucial importance of the DA information for assessing dialogue coherence.

\section{State of the art}
\label{sec:soa}
\begin{figure*}[]
%\centering
%\subcaption*{A: Entity grid}
\begin{tabular}{l|lllll}
%\begin{tabular}{l|ll>{\color{red}}llllllllll}
   & \rotatebox{90}{company} & \rotatebox{90}{drugs} & \rotatebox{90}{policy} & \rotatebox{90}{convictions} & \rotatebox{90}{clients} \\
  \hline
\textbf{t1} & S & X & - & - & -\\
\textbf{t2} & X & S & O & X & S\\
\textbf{t3} & - & - & - & - & -\\
\textbf{t4} & - & - & - & - & -\\
\textbf{t5} & - & X & - & - & -\\
& & & & &\\
& & & & &\\
& & & & &\\
\multicolumn{6}{c}{A: \textit{Entity Grid}}\\
% \textbf{t5} & S & O& \leavevmode\color{red}S & X & O 
\end{tabular} %\subcaption*{B: Modified grid}
 \begin{tabular}{l|llllll}
%\begin{tabular}{l|ll>{\color{red}}llllllllll}
   & \rotatebox{90}{company} & \rotatebox{90}{drugs} & \rotatebox{90}{policy} & \rotatebox{90}{convictions} & \rotatebox{90}{clients} & \rotatebox{90}{\textit{no\_entities}} \\
  \hline
\textbf{da1} & qy & qy & - & - & - & -\\
\textbf{da2} & - & na & na & - & - & -\\
\textbf{da3} & - & - & - & - & - & sd\^e\\
\textbf{da4} & sd & sd & sd & sd & sd & -\\
\textbf{da5} &  - & - & - & - & - & \%\\
\textbf{da6} & - & - & - & - & - & qo\\
\textbf{da7} & - & - & - & - & - & nn\\
\textbf{da8} & - & sd\^e & - & - & - & -\\
\multicolumn{7}{c}{B: \textit{Modified Grid}}\\
% \textbf{t5} & S & O& \leavevmode\color{red}S & X & O 
\end{tabular}
% \vspace{1mm} \\
 \scalebox{0.9}{
\begin{minipage}[]{0.44\textwidth}
\begin{flushleft}
\small{
\textbf{\textit{A.} t.1 da.qy} Well does \lbrack the company \rbrack \textsubscript{S} you work for test for\lbrack drugs\rbrack \textsubscript{X}?\\
\textbf{\textit{B.} t.2 da.na} Actually, they just recently started \lbrack a policy\rbrack \textsubscript{O} of testing \lbrack drugs\rbrack \textsubscript{X}, which was kind of interesting,\\
\textbf{\textit{B.} t.2 da.sd\^e} because when I went to work for them, uh, they didn't do that\\
\textbf{\textit{B.} t.2 da.sd} but, uh, since then they've started a \lbrack drug\rbrack \textsubscript{O} testing \lbrack policy\rbrack \textsubscript{O}, not because of their own, uh \lbrack convictions\rbrack \textsubscript{X}, but because \lbrack the clients\rbrack \textsubscript{S} of \lbrack our company\rbrack \textsubscript{X} are requests that we do that.\\
\textbf{\textit{A.} t.3 da.\%} Huh.\\
\textbf{\textit{B.} t.4 da.qo} How about you?\\
\textbf{\textit{A.} t.5 da.nn} Uh, no\\
\textbf{\textit{A.} t.5 da.sd\^e} we're not being tested for \lbrack drugs\rbrack \textsubscript{X} at all, uh \\
}
\end{flushleft}
\end{minipage}
}
\caption{Entity grid example (A) vs. our modified grid (B) for the extract of the Switchboard dialogue 002\_4330 (on the right). Entities in the sentences are annotated with their syntactic role: subject (S), object (O) or neither (X). The Dialogue Act tags are directly taken from the SWBD DAMSL annotation: qy (yes-no-question), na (affirmative-non-yes-answers), sde (Statement expanding y/n answer), sd (statement-non-opinion), \% (uninterpretable), qo (open-question), nn (no-answers).}
% DAs: qy (yes-no-question), na (affirmative-non-yes-answers), sde (Statement expanding y/n answer), sd (statement-non-opinion), \%(uninterpretable), qo(open-question), nn(no-answers)
\label{tab:entitygrid}
\end{figure*}

The most fertile framework for local coherence modelling in text is arguably the \textit{entity grid} \cite{barzilay2008modeling}.
As shown in Figure~\ref{tab:entitygrid}, this approach proposes to represent the structure of a document (in our case a dialogue) through a grid displaying transitions in the syntactic roles of entities (the heads of Noun Phrases (NP)) across neighbouring sentences in the text. In the grid, the rows represent subsequent sentences (turns in our case, as in \cite{elsner2011disentangling}) while each entity is represented by a column. A grammatical role can be: subject ($S$), direct object ($O$) or neither ($X$), plus a symbol ($-$) to signal that an entity does not appear in that turn $t$.
The assumption is that the grid topology of coherent texts exhibits certain regularities associated to the way entities are  introduced and become the focus of the discourse.
For example, in the case of the grid represented in Figure~\ref{tab:entitygrid}, Table A we can notice how the sentences are connected by the continuity of the entity ``drugs'' across different turns. If an entity appears more than once in the same turn the most prominent syntactic role is chosen ($S>O>X$). %We notice that while the original approach considers only the head of the NPs, we follow 

By computing the probabilities over all possible transitions of length \textit{n} from one category to all others (thus $\left\{S,O,X,-\right\}^{n}$) we can turn this representation into a feature vector, similar to a language model over the entity tags, representing the syntactic role transitions of entities in the whole document. It is important to notice that the entity grid is not lexicalised, since this information is lost when creating the feature vectors.

In \cite{barzilay2008modeling}, the authors use these feature vectors to train a Support Vector Machine (using SVM\textsuperscript{light} \cite{joachims2002optimizing}) modelling coherence as a ranking problem and using as a training dataset a set of original documents as positive (coherent) examples, paired with a set of the same documents with the sentences randomly permuted as negative (incoherent) examples (a procedure called pairwise training). The authors also experiment with models using different degrees of saliency (entity frequency) and transitions lengths (between 2 and 4), and by employing coreference resolution systems to detect entity chains (however given the performances of SoA systems this addition does not provide improvements).

The algorithm proposed in \cite{barzilay2008modeling} derives thus automatically an abstract representation for a text, with as the only requirement a syntactic parser and a dataset. Among the weak points of this framework, however, is the fact that it models only local coherence (patterns of distribution across adjacent sentences) and a data sparsity problem.

%Entity grids extensions
Over the years, the entity grid model inspired numerous extensions \cite{guinaudeau2013graph, filippova2007extending, elsner2011extending} and similar implementations. Some approaches \cite{filippova2007extending}, for example, augmented the model using the semantic relatedness of the entities but without much improvement. Others \cite{elsner2011extending} showed the usefulness of incorporating entity--specific features such as named entity information and considering also nouns which do not head NPs (as in Figure~\ref{tab:entitygrid}, turn 2, where in the NP ``a drug testing policy'' we consider both ``drug'' and ``policy'' as entities).
 %extended the entity grid by adding all nouns (not just head nouns as in the original approach\cite{barzilay2008modeling}) to the grid showing consistent improvement and . 

%The features for coreference are picked up using in-domain data, so that it is not required to run coreference during testing.
%Intuition behind this work: some entities are more likely to be prominent for the text genre under test (here news articles, they expect people and organizations to be more prominent than dates). Best results.  
The typical tasks on which local coherence models are currently evaluated are: sentence ordering \textit{discrimination}, where the system needs to rank original documents higher than randomly permuted ones, and \textit{insertion}, introduced by \cite{elsner2011extending}, where the system has to rank the position of a sentence removed from a document. 
%In the insertion task the system has to locate the position of a sentence removed from a document. %Add mention to summary coherence rating? 
The state of the art for these tasks was recently achieved by \cite{nguyen2017neural}, which uses the entity grid as input to a Convolutional Neural Network. The authors report an accuracy of 88.69 (compared to 81.58 of the original grid with both head and non-head nouns) and an insertion score of 25.95 (compared to 22.13 of the same model). One of the advantages of the neural model compared to the original one is its ability to model long range entity transitions. Other recent works inspired by the entity grid include coherent paragraph generation \cite{li2017neural}, and applications to automated essay scoring \cite{farag2018neural} and neural stories text generation \cite{clark2018neural}.

% @Alessandra: apply or transfer?
%Importantly, it has been shown that 
Entity-based local coherence models apply well to dialogue as is or with some extra features, but not DAs
%without much modifications by testing the model
%on Switchboard 
\cite{purandare2008analyzing,elsner2011disentangling}.
%Besides the studies that transferred the entity grid to dialogue, other approaches to
Dialogue coherence has been explored outside of the entity grid approach as well
%Other studies explored dialogue coherence assessment independently from the entity grid literature 
\cite{higashinaka2014evaluating,gandhe2016semi,venkateshevaluating}. In \cite{gandhe2016semi}, the authors propose a semi-automatic approach to evaluate dialogue coherence using only DA and relying on turn level coherence ratings from multiple sources. To the best of our knowledge, the only approach that combines entity and DA information for dialogue coherence evaluation is \cite{higashinaka2014evaluating}, which did not utilize the entity grid and models coherence as a binary classification task on utterance pairs rather than the whole conversation.
%Unlike \cite{higashinaka2014evaluating}, who also utilized DA and entity information, but not as an entity grid % only utterance pairs
%\textcolor{red}{Contrast dialogue coherence papers: metrics never compared to coherence tasks, \cite{gandhe2016semi} it is semiautomatic: it relies on contrasting several judgments and only on DAs without entities,   \cite{higashinaka2014evaluating} also combines DAs and entities, but uses utterance pairs without previous context and no theory about entities transitions}

%final remarks on entity grid
% All of these models, however, are only aimed at capturing local coherence, that is the surface features that make a text locally connected.
\begin{table}
\centering
\begin{tabularx}{\columnwidth}{|l|R|R|R|}
\hline 
& \multicolumn{1}{c|}{\textbf{SWBD}}
& \multicolumn{1}{c|}{\textbf{AMI}} % 1-2 3-5
& \multicolumn{1}{c|}{\textbf{Oasis}} \\
\hline\hline
%\multicolumn{3}{|c|}{\textbf{Length and Sentiment}}\\
%\hline
\# DA tags  &  42 & 16 & 41 \\
Av. \# tokens/turn &  13.1 & 15.1 & 10.6 \\
Av. \# turns/dialogue &  109 & 86.4 & 10.6 \\
%\hline\hline
%\multicolumn{3}{|c|}{\textbf{Sentiment}}\\
%\hline
\hline
\# Train dialogues &  740 & 356 & 191 \\
\# Test dialogues &  231 & 111 & 59 \\
\# Dev dialogues &  184 & 89 & 45 \\
%... & & \\
\hline
\end{tabularx}
\caption{We report the count of Dialogue Act tags, the average number of tokens per turn, the average count of turns per dialogue and counts of our Training, test and Developments splits for our three datasets: SWBD, AMI and Oasis. The document counts are given for the original documents, therefore need to be multiplied times 20 (pairs) for the discrimination task and times 100 (10x10) for the insertion task.}
\label{tbl:corr}
\end{table}
\section{Methodology}
\label{sec:methodology}
Both the original and its SoA extensions for coherence assessment focus on modelling local (entity-based) coherence, which is a form of surface coherence of the text (cohesion in Pragmatics theory \cite{halliday1976cohesion}). However we can easily imagine how the entity grid or its extensions would not capture the lack of coherence in the following example:\\\\
A. Do you have dogs?\\
B. What is the average height of dogs?\\\\
In this case the text would be judged coherent given the continuation of the entity ``dogs'' across both turns. Nonetheless this example is incoherent because B does not answer A's question, but rather introduces an unrelated question.

In this work we augment the original entity grid document representation with a notion of global coherence, as provided by the intentional structure of the conversation in the form of Dialogue Acts. Our hypothesis is that DA information could improve coherence models performance on dialogue. This hypothesis is also motivated by the fact that syntactic roles might no be so prominent or reliable when transferred to the spoken dialogue domain, since for some dialogue types turns tend to be quite short and syntactic parsers are not very robust when there is no punctuation.

In order to test our hypothesis, we experiment with various grid constructions in order to find the best way to combine the DAs information with the original representation. 
For clarity, we follow a template $<$row$>$-Grid:$<$cell$>$ for naming our different document representations. In particular the $<$row$>$ refers to text span (row in the grid) chosen, either the Turn (T) as in \cite{elsner2011disentangling} or the text span of the DA (D); the $<$cell$>$ refers to the category in the grid cells, either the syntactic role (\textit{role}), the presence of the entity (\textit{presence}, reducing the vocabulary to entities presence ($X$) or not ($-$) already proposed in \cite{barzilay2008modeling}) or the DA tag (\textit{DA}, which varies according the DA schema of each dataset). In the rest of the section we detail the document representations in our experiments. \\\\
\textbf{Baselines}: The baselines \textit{T-Grid:roles} and \textit{T-Grid:presence} replicate respectively the original entity grid in its all nouns variant (proposed by \cite{elsner2011extending}) and a simplified version of the grid where the vocabulary is restricted to two items.\\
\textbf{D-Grid:role}: This variation differs from the \textit{T-Grid:roles} only for the fact that the text span units are DAs, rather than turns, while the vocabulary is still composed by syntactic roles. The disadvantage of this representation is that it is more sparse than its preceding one, but it is able to capture in-turn entities transitions. \\
\textbf{D-Grid:DA}: In this variant the syntactic roles tags are substituted by the DA categories (according to the dataset's DA scheme). This is the modified grid shown in Figure~\ref{tab:entitygrid}, Table B. In this document representation an extra "no\_entities" column is added to capture the DA tags where no entity is mentioned. \\ 
\textbf{Only DAs}: This text representation is the same as the previous one, with the difference that here all entities are dropped and we keep only one column with all the DAs.\\
\textbf{Combinations}: \textit{T-Grid:presence + Only DAs} and \textit{T-Grid:role + Only DAs} represent the combination of \textit{Only DAs} with the two baselines by simply concatenating their feature vectors. These variations combine the entities and DAs feature vectors as two separate sources of information. 
\begin{table*}
\setlength{\tabcolsep}{4pt}
\centering
%\begin{small}
\begin{tabularx}{\textwidth}{l|RRR|R|RRR|R|RRR|R|}
\cline{2-13}
& \multicolumn{4}{c|}{\textbf{SWBD}}
& \multicolumn{4}{c|}{\textbf{AMI}}
& \multicolumn{4}{c|}{\textbf{Oasis}}\\
\cline{2-13}
& \multicolumn{3}{c|}{\textbf{Discr.}}
& \multicolumn{1}{c|}{\textbf{Ins.}}
& \multicolumn{3}{c|}{\textbf{Discr.}}
& \multicolumn{1}{c|}{\textbf{Ins.}}
& \multicolumn{3}{c|}{\textbf{Discr.}}
& \multicolumn{1}{c|}{\textbf{Ins.}}\\
% \cline{2-13}
& \multicolumn{1}{c}{\textbf{Acc.}}
& \multicolumn{1}{c}{\textbf{MRR}}
& \multicolumn{1}{c|}{\textbf{P@1}}
& \multicolumn{1}{c|}{\textbf{Av. P@1}}
& \multicolumn{1}{c}{\textbf{Acc.}}
& \multicolumn{1}{c}{\textbf{MRR}}
& \multicolumn{1}{c|}{\textbf{P@1}}
& \multicolumn{1}{c|}{\textbf{Av. P@1}}
& \multicolumn{1}{c}{\textbf{Acc.}}
& \multicolumn{1}{c}{\textbf{MRR}}
& \multicolumn{1}{c|}{\textbf{P@1}}
& \multicolumn{1}{c|}{\textbf{Av. P@1}}\\
\hline
Random 
& 50.00 & 16.98 &  4.76 &  8.70 
& 50.00 & 18.93 &  6.31 &  9.44
& 50.00 & 17.39 &  5.08 &  9.16\\
Only DAs %D-Grid: 
& \textbf{99.76} & 98.76 & 97.80 & 45.45
& \textbf{98.78} & \textbf{95.27} & \textbf{92.79} & 30.75
& 91.53 & 68.47 & 54.24 & 41.44\\
\hline
T-Grid:presence % T-Grid:da
& 70.65 & 38.60 & 24.24 & 10.74
& 76.71 & 40.88 & 25.23 &  7.21
& 72.03 & 33.94 & 18.64 & 23.49\\
T-Grid:role 
& 64.78 & 29.39 & 13.85 & 12.08
& 79.59 & 46.73 & 28.83 & 11.71
& 65.25 & 26.34 & 10.17 & 18.80\\
D-Grid:role 
& 63.25 & 28.50 & 13.85 & 10.00
& 59.41 & 25.40 & 11.71 & 11.71
& 49.58 & 17.08 &  3.39 & 15.52\\
\hline
D-Grid:DA &
99.57 & 97.36 & 95.67 & 38.79
& 95.41 & 83.02 & 75.68 & 19.25
& 87.80 & 57.64 & 40.68 & 28.96\\
\hline
T-Grid:presence + Only DAs 
& \textbf{99.76} & 98.76 & 97.84 & \textbf{45.58}
& 98.47 & 93.74 & 90.09 & 31.41
& \textbf{92.46} & 69.75 & \textbf{57.63} & \textbf{42.74}\\
T-Grid:role + Only DAs 
& 99.68 & \textbf{99.17} & \textbf{98.70} & 44.98 
& 98.51 & 94.56 & 91.89 & \textbf{32.43}
& 91.78 & \textbf{70.39} & \textbf{57.63} & 42.49\\
\hline
\end{tabularx}
%\end{small}
\caption{For each of the three datasets considered (SWBD, AMI and Oasis) we report results on the two tasks of Discrimination and Insertion. For Discrimination, we report the standard Accuracy (Acc.), plus Mean Reciprocal Rank (MRR) and Precision at one (P@1). For Insertion, we report the standard metric for this task, i.e. Precision at one (P@1) averaged for the dialogue.}
\label{tbl:pred}
\end{table*}

\section{Experimental setup}
\label{sec:experiments}
\textbf{Tasks}: We evaluate our models on the sentence ordering \textit{discrimination} task proposed in the original \cite{barzilay2008modeling} and on the \textit{insertion} task proposed in \cite{elsner2011extending}, which represent the standard evaluation tasks for coherence models. In order to ensure comparability across our experiments, when permuting the order in the documents, we always permute the entire turn (therefore multiple rows in case we have several DAs in the same turn) and the same permutations are kept across all settings.

The first task, discrimination, is usually evaluated as accuracy of the model in ranking the original text higher than a permuted one (we use 20 permutations per document following previous work \cite{barzilay2008modeling,elsner2011extending,nguyen2017neural}). In order to better analyse our results, we add to this metric two widely used ranking metrics, i.d. Mean Reciprocal Rank (MRR, the average of reciprocal ranks in a set of queries) and Precision at One (P@1, the ability of the model to rank the original higher than all the permutations). In both these metrics, instead of comparing the original document with each of its permutation we compare the rank of the original document to all its permutations at the same time.

On the other hand, the insertion task is evaluated as the average number of sentences per document inserted in the correct position (therefore the average of the P@1). For the insertion task, we randomly pick 10 turns per dialogue and insert each one in 10 random positions (for each dataset we used the same turns and positions to ensure intra-dataset comparability). 
\\
\textbf{Datasets}: In order to verify the robustness of our models across different DAs schemes and dialogue types, we perform all our experiments on three different publicly available datasets with DA annotation, namely BT Oasis\cite{leech2003generic}, AMI\cite{amicorpus} and the Switchboard Dialogue Act corpus \cite{godfrey1992switchboard} (SWBD). Table~\ref{tbl:corr} shows some differences across the datasets.\footnote{The code is available at: \small{\url{https://github.com/alecervi/Coherence-models-for-dialogue}}}

The dialogues in SWBD are open-domain telephone conversations. The individual turns tend to be quite long while the dialogues are the longest across the three datasets. For the DA categories we employ the 42 DAMSL ones.
Oasis, on the other hand, is quite the opposite. A dataset of task-based conversations between clients and British Telecom help desk, here the turns tend to be quite short and the dialogues very short. 
AMI presents yet another type of dialogue data. Compared to the other datasets here the dialogues are between multiple speakers. In these dialogues participants were asked to discuss a project, so turns tend to be very long. This is also the dataset with the less rich annotation scheme compared to the previous two (only 16 DA categories).
\\
\textbf{Parameters}: As in the original entity grid paper we test all our models using the preference kernel implemented in SVM\textsuperscript{light} \cite{joachims2002optimizing} with default parameters. We follow the default original grid parameters (saliency:1, transitions length:2) for all our experiments. This was done to ensure a fair comparison between the datasets with few entities and short dialogues (Oasis) and those with many turns and several entities (Switchboard, AMI).
For preprocessing the text to extract noun phrases and their syntactic roles we use spacy \cite{honnibal-johnson:2015:EMNLP}.

\section{Results}
\label{sec:results}
We report the results of our experiments in Table~\ref{tbl:pred}.
To the model described in Section ~\ref{sec:methodology} we add a Random baseline, to give a measure of how the difficulty of both tasks vary across the datasets.
To assess the respective significance of the coherence models, for discrimination accuracy and P@1 we use the McNemar test, while for discrimination MRR and the insertion Average P@1 we use Fisher's randomization test.%; both with $p = 0.05$.

Regarding the \textit{discrimination} task, the first thing to notice is how Only DAs, the model capturing DAs transitions without taking into account entities information, is a very competitive model across all the three datasets. Indeed the intentional structure information alone is so strong that on SWBD and AMI, the  task of discriminating an original document from randomly shuffled re-orderings of the same document seems even too easy. With similar setup and data (also Switchboard but a different subset of dialogues with 505 original dialogues for training and 153 for testing) \cite{elsner2011disentangling} reports an accuracy of 86.0 for its extended version of the grid.
The strength of the intentional structure information is still prominent, but less visible in Oasis, where the dialogues are much shorter compared to the previous two datasets and it might be possible that random shuffling of turns might not disrupt the dialogue coherence so effectively.

In general, we notice the importance of DA information across the three datasets also for the rest of the proposed models for the discrimination task. As expected, the lowest results are achieved by the D-Grid:role model, which are still much better than the Random baseline. This model is similar to the original grid with the disadvantage of increasing the sparsity of entities.

The next lowest scores are then achieved by the T-Grid:presence and T-Grid:role. While the second performs better on AMI, where turns are the longest and we can expect sentence structure to be more complicated, the T-Grid:presence outperforms T-Grid:role both on Oasis and SWBD, confirming our hypothesis regarding the diminished importance of syntactic roles in dialogue.
The next best model across all datasets for discrimination is D-Grid:DA with a large distance compared to T-Grid:presence and T-Grid:role.

The best performing models are the combinations, where the entity and DA information are encoded separately. These models achieve the best results on SWBD and Oasis, while their distance to Only DAs is not statistically significant on AMI.

The observations made on the discrimination task are reinforced on the \textit{insertion} task. Only by looking at the performances (between 8.70 and 9.44) for the Random baseline, we notice how much the task is harder than the previous one (as mentioned in \ref{sec:soa} the SOA in the Wall Street Journal is 25.95). 
The noticeable difference in the results for the T-Grid:presence, T-Grid:role compared to D-Grid:DA for insertion confirms once again how crucial is the intentional information.
While also for insertion the intentional structure alone gives a very strong signal across all the datasets, the best results are achieved by combining the DAs with the entity information. This result is consistent with the nature of the task, where entity information could provide an important contribution to locating the exact place of a turn in the conversation.
Also for this task, the syntactic role information yields the highest scores only for AMI, the dataset with the longest turns, while on SWBD and Oasis the best results are achieved by the simpler model -- T-Grid:presence + Only DAs. 

The Only-DAa model significantly outperforms the entity grid coherence models without DAs. However, while the models using the combination of entity grid and DAs (T-Grid:presence + Only DAs, T-Grid:role + Only DAs) yield better performance on SWDA and Oasis, overall their differences are not statistically significant.

\section{Conclusions}
\label{sec:conclusions}
In this paper, we applied the entity grid local coherence approach to dialogue. We experimented with different variations of its document representation in order to find the best way to augment it with participants' intents, an expression of global coherence and a signal which has been widely studied in dialogue to describe the structure of conversations. Our experiments confirm the crucial importance of the intentional structure for dialogue coherence, but also show how its combination with entity information could be useful for harder tasks connected to dialogue coherence, such as insertion. 

Furthermore, our experiments show how the task of sentence ordering discrimination might be too easy on dialogue, where the DAs already give a very strong signal. On the other hand, the task of insertion is by far more difficult. For future work, we plan to explore other tasks for coherence modelling that might be more useful for dialogue, such as automatic prediction of the next dialogue turn.

It is also important to notice that our proposals for document representation are independent of the Machine Learning models employed. They could therefore be used, for example, in combination with a CNN as implemented in \cite{nguyen2017neural}. Another application we foresee for these models is to be used in the reward function for training dialogue systems in a Reinforcement Learning setting. 
Moreover, it is worth noticing that our experiments were performed using gold DAs. One of the first future experiments to perform would be to replicate the experiments with predicted DA labels, rather than gold ones to verify the robustness of the approach when using a DA tagger (the current approaches to DA tagging on Switchboard report accuracies above 75\% \cite{ji2016latent,mezza2018iso}). In such a setting, we imagine that the entities information might play even more important role in assessing dialogue coherence.
Other possible directions include applying our coherence models to chat disentanglement, as well as the automatic evaluation of conversational agents' coherence.

\newpage
\bibliography{strings}

% Generated by IEEEtran.bst, version: 1.13 (2008/09/30)
\begin{thebibliography}{10}
\providecommand{\url}[1]{#1}
\csname url@samestyle\endcsname
\providecommand{\newblock}{\relax}
\providecommand{\bibinfo}[2]{#2}
\providecommand{\BIBentrySTDinterwordspacing}{\spaceskip=0pt\relax}
\providecommand{\BIBentryALTinterwordstretchfactor}{4}
\providecommand{\BIBentryALTinterwordspacing}{\spaceskip=\fontdimen2\font plus
\BIBentryALTinterwordstretchfactor\fontdimen3\font minus
  \fontdimen4\font\relax}
\providecommand{\BIBforeignlanguage}[2]{{%
\expandafter\ifx\csname l@#1\endcsname\relax
\typeout{** WARNING: IEEEtran.bst: No hyphenation pattern has been}%
\typeout{** loaded for the language `#1'. Using the pattern for}%
\typeout{** the default language instead.}%
\else
\language=\csname l@#1\endcsname
\fi
#2}}
\providecommand{\BIBdecl}{\relax}
\BIBdecl

\bibitem{li2016diversity}
J.~Li, M.~Galley, C.~Brockett, J.~Gao, and B.~Dolan, ``A diversity-promoting
  objective function for neural conversation models,'' in \emph{Proceedings of
  NAACL-HLT}, 2016, pp. 110--119.

\bibitem{li2016deep}
J.~Li, W.~Monroe, A.~Ritter, D.~Jurafsky, M.~Galley, and J.~Gao, ``Deep
  reinforcement learning for dialogue generation,'' in \emph{Proceedings of the
  2016 Conference on Empirical Methods in Natural Language Processing}, 2016,
  pp. 1192--1202.

\bibitem{liu2016not}
C.-W. Liu, R.~Lowe, I.~Serban, M.~Noseworthy, L.~Charlin, and J.~Pineau, ``How
  not to evaluate your dialogue system: An empirical study of unsupervised
  evaluation metrics for dialogue response generation,'' in \emph{Proceedings
  of the 2016 Conference on Empirical Methods in Natural Language Processing},
  2016, pp. 2122--2132.

\bibitem{papineni2002bleu}
K.~Papineni, S.~Roukos, T.~Ward, and W.-J. Zhu, ``Bleu: a method for automatic
  evaluation of machine translation,'' in \emph{Proceedings of the 40th annual
  meeting of the Association for Computational Linguistics}.\hskip 1em plus
  0.5em minus 0.4em\relax Association for Computational Linguistics, 2002, pp.
  311--318.

\bibitem{lowe2016evaluation}
R.~Lowe, I.~V. Serban, M.~Noseworthy, L.~Charlin, and J.~Pineau, ``On the
  evaluation of dialogue systems with next utterance classification,'' in
  \emph{17th Annual Meeting of the Special Interest Group on Discourse and
  Dialogue}, 2016, p. 264.

\bibitem{walker1997paradise}
M.~A. Walker, D.~J. Litman, C.~A. Kamm, and A.~Abella, ``Paradise: A framework
  for evaluating spoken dialogue agents,'' in \emph{Proceedings of the eighth
  conference on European chapter of the Association for Computational
  Linguistics}.\hskip 1em plus 0.5em minus 0.4em\relax Association for
  Computational Linguistics, 1997, pp. 271--280.

\bibitem{gandhe2016semi}
S.~Gandhe and D.~Traum, ``A semi-automated evaluation metric for dialogue model
  coherence,'' \emph{Situated Dialog in Speech-Based Human-Computer
  Interaction}, p. 217, 2016.

\bibitem{higashinaka2014evaluating}
R.~Higashinaka, T.~Meguro, K.~Imamura, H.~Sugiyama, T.~Makino, and Y.~Matsuo,
  ``Evaluating coherence in open domain conversational systems,'' in
  \emph{Fifteenth Annual Conference of the International Speech Communication
  Association}, 2014.

\bibitem{venkateshevaluating}
A.~Venkatesh, C.~Khatri, A.~Ram, F.~Guo, R.~Gabriel, A.~Nagar, R.~Prasad,
  M.~Cheng, B.~Hedayatnia, A.~Metallinou, R.~Goel, S.~Yang, and A.~Raju, ``On
  evaluating and comparing conversational agents,'' in \emph{NIPS 2017
  Conversational AI workshop}, 2017.

\bibitem{grosz1995centering}
B.~J. Grosz, S.~Weinstein, and A.~K. Joshi, ``Centering: A framework for
  modeling the local coherence of discourse,'' \emph{Computational
  linguistics}, vol.~21, no.~2, pp. 203--225, 1995.

\bibitem{barzilay2008modeling}
R.~Barzilay and M.~Lapata, ``Modeling local coherence: An entity-based
  approach,'' \emph{Computational Linguistics}, vol.~34, no.~1, pp. 1--34,
  2008.

\bibitem{joshi1979centered}
A.~K. Joshi and S.~Kuhn, ``Centered logic: The role of entity centered sentence
  representation in natural language inferencing.'' in \emph{IJCAI}, 1979, pp.
  435--439.

\bibitem{givon1987beyond}
T.~Giv{\'o}n, ``Beyond foreground and background,'' \emph{Coherence and
  grounding in discourse}, vol.~11, pp. 175--188, 1987.

\bibitem{elsner2011extending}
M.~Elsner and E.~Charniak, ``Extending the entity grid with entity-specific
  features,'' in \emph{Proceedings of the 49th Annual Meeting of the
  Association for Computational Linguistics: Human Language Technologies: short
  papers-Volume 2}.\hskip 1em plus 0.5em minus 0.4em\relax Association for
  Computational Linguistics, 2011, pp. 125--129.

\bibitem{purandare2008analyzing}
A.~Purandare and D.~J. Litman, ``Analyzing dialog coherence using transition
  patterns in lexical and semantic features.'' in \emph{FLAIRS Conference},
  2008, pp. 195--200.

\bibitem{elsner2011disentangling}
M.~Elsner and E.~Charniak, ``Disentangling chat with local coherence models,''
  in \emph{Proceedings of the 49th Annual Meeting of the Association for
  Computational Linguistics: Human Language Technologies}, 2011, pp.
  1179--1189.

\bibitem{sacks1974simplest}
H.~Sacks, E.~A. Schegloff, and G.~Jefferson, ``A simplest systematics for the
  organization of turn-taking for conversation,'' \emph{language}, pp.
  696--735, 1974.

\bibitem{sacks1995lectures}
H.~Sacks and G.~Jefferson, ``Lectures on conversation,'' 1995.

\bibitem{schegloff1968sequencing}
E.~A. Schegloff, ``Sequencing in conversational openings,'' \emph{American
  anthropologist}, vol.~70, no.~6, pp. 1075--1095, 1968.

\bibitem{schegloff1973opening}
E.~A. Schegloff and H.~Sacks, ``Opening up closings,'' \emph{Semiotica},
  vol.~8, no.~4, pp. 289--327, 1973.

\bibitem{austin1975things}
J.~L. Austin, \emph{How to do things with words}.\hskip 1em plus 0.5em minus
  0.4em\relax Oxford university press, 1975.

\bibitem{grosz1986attention}
B.~J. Grosz and C.~L. Sidner, ``Attention, intentions, and the structure of
  discourse,'' \emph{Computational linguistics}, vol.~12, no.~3, pp. 175--204,
  1986.

\bibitem{allen1980analyzing}
J.~F. Allen and C.~R. Perrault, ``Analyzing intention in utterances,''
  \emph{Artificial intelligence}, vol.~15, no.~3, pp. 143--178, 1980.

\bibitem{godfrey1992switchboard}
J.~J. Godfrey, E.~C. Holliman, and J.~McDaniel, ``Switchboard: Telephone speech
  corpus for research and development,'' in \emph{Acoustics, Speech, and Signal
  Processing, 1992. ICASSP-92., 1992 IEEE International Conference on},
  vol.~1.\hskip 1em plus 0.5em minus 0.4em\relax IEEE, 1992, pp. 517--520.

\bibitem{amicorpus}
J.~Carletta, ``Announcing the ami meeting corpus,'' \emph{The ELRA Newsletter
  11(1), January-March, p. 3-5.}, 2006.

\bibitem{leech2003generic}
G.~Leech and M.~Weisser, ``Generic speech act annotation for task-oriented
  dialogues,'' in \emph{Procs. of the 2003 Corpus Linguistics Conference, pp.
  441Y446. Centre for Computer Corpus Research on Language Technical Papers,
  Lancaster University}, 2003.

\bibitem{joachims2002optimizing}
T.~Joachims, ``Optimizing search engines using clickthrough data,'' in
  \emph{Proceedings of the eighth ACM SIGKDD international conference on
  Knowledge discovery and data mining}.\hskip 1em plus 0.5em minus 0.4em\relax
  ACM, 2002, pp. 133--142.

\bibitem{guinaudeau2013graph}
C.~Guinaudeau and M.~Strube, ``Graph-based local coherence modeling.'' in
  \emph{ACL (1)}, 2013, pp. 93--103.

\bibitem{filippova2007extending}
K.~Filippova and M.~Strube, ``Extending the entity-grid coherence model to
  semantically related entities,'' in \emph{Proceedings of the Eleventh
  European Workshop on Natural Language Generation}.\hskip 1em plus 0.5em minus
  0.4em\relax Association for Computational Linguistics, 2007, pp. 139--142.

\bibitem{nguyen2017neural}
D.~T. Nguyen and S.~Joty, ``A neural local coherence model,'' in
  \emph{Proceedings of the 55th Annual Meeting of the Association for
  Computational Linguistics (Volume 1: Long Papers)}, vol.~1, 2017, pp.
  1320--1330.

\bibitem{li2017neural}
J.~Li and D.~Jurafsky, ``Neural net models of open-domain discourse
  coherence,'' in \emph{Proceedings of the 2017 Conference on Empirical Methods
  in Natural Language Processing}, 2017, pp. 198--209.

\bibitem{farag2018neural}
Y.~Farag, H.~Yannakoudakis, and T.~Briscoe, ``Neural automated essay scoring
  and coherence modeling for adversarially crafted input,'' in
  \emph{Proceedings of the 2018 Conference of the North American Chapter of the
  Association for Computational Linguistics: Human Language Technologies,
  Volume 1 (Long Papers)}, vol.~1, 2018, pp. 263--271.

\bibitem{clark2018neural}
E.~Clark, Y.~Ji, and N.~A. Smith, ``Neural text generation in stories using
  entity representations as context,'' in \emph{Proceedings of the 2018
  Conference of the North American Chapter of the Association for Computational
  Linguistics: Human Language Technologies, Volume 1 (Long Papers)}, vol.~1,
  2018, pp. 2250--2260.

\bibitem{halliday1976cohesion}
``Cohesion in english, author={Halliday, Michael Alexander Kirkwood and Hasan,
  Ruqaiya}, year={1976}, publisher={Routledge}.''

\bibitem{honnibal-johnson:2015:EMNLP}
\BIBentryALTinterwordspacing
M.~Honnibal and M.~Johnson, ``An improved non-monotonic transition system for
  dependency parsing,'' in \emph{Proceedings of the 2015 Conference on
  Empirical Methods in Natural Language Processing}.\hskip 1em plus 0.5em minus
  0.4em\relax Lisbon, Portugal: Association for Computational Linguistics,
  September 2015, pp. 1373--1378. [Online]. Available:
  \url{https://aclweb.org/anthology/D/D15/D15-1162}
\BIBentrySTDinterwordspacing

\bibitem{ji2016latent}
Y.~Ji, G.~Haffari, and J.~Eisenstein, ``A latent variable recurrent neural
  network for discourse relation language models,'' in \emph{Proceedings of
  NAACL-HLT}, 2016, pp. 332--342.

\bibitem{mezza2018iso}
S.~Mezza, A.~Cervone, G.~Tortoreto, E.~A. Stepanov, and G.~Riccardi,
  ``{ISO}-standard domain-independent dialogue act tagging for conversational
  agents,'' in \emph{COLING}, 2018.

\end{thebibliography}
\bibliographystyle{IEEEtran}
\end{document}